\documentclass[twoside,11pt]{article}
\usepackage{jmlr2e_copy_11_6_18}
\usepackage{subcaption}

\jmlrheading{82}{2017}{1-9}{}{K. Gretchen Greene}{4th International Conference on Predictive Applications and APIs}

\ShortHeadings{Rule Based Bootstrapping for Small Data}{Greene}

\begin{document}

\title{DragonPaint: Rule Based Bootstrapping for Small Data with an Application to Cartoon Coloring}

\author{\name K. Gretchen Greene \email kgretchengreene@gmail.com \\
	\addr Greene Analytics\\
	Industry Lab\\
	288 Norfolk St.\\
	Cambridge, MA 02139, USA}

\firstpageno{1}

\maketitle

\begin{abstract}%
In this paper, we confront the problem of deep learning's big labeled data requirements, offer a rule based strategy for extreme augmentation of small data sets and apply that strategy with the image to image translation model by \citet{pix2pix:16} to automate cel style cartoon coloring with very limited training data.  While our experimental results using geometric rules and transformations demonstrate the performance of our methods on an image translation task with industry applications in art, design and animation, we also propose the use of rules on partial data sets as a generalizable small data strategy, potentially applicable across data types and domains.
\end{abstract}

\begin{keywords}
	rules, augmentation, image to image translation, style transfer, synthetic data 
\end{keywords}

\section{Introduction}

A big problem with supervised machine learning is the need for huge amounts of labeled data. At least it's a big problem if you don't have the labeled data and even now, in a world awash with big data, most of us don't. While a few companies have access to enormous quantities of certain kinds of labeled data, for most organizations and many applications, the creation of sufficient quantities of exactly the kind of labeled data desired is cost prohibitive or impossible. Sometimes the domain is one where there just isn't much data. It might be the diagnosis of a rare disease or determining whether a signature matches a few known exemplars. Other times the volume of data needed and the cost of human labeling by Amazon Turkers or summer interns is just too high. Paying to label every frame of a movie length video adds up fast, even at a penny a frame.

We confront the problem of deep learning's big labeled data requirements, offer a rule based strategy for extreme augmentation of small data sets and apply that strategy with an image to image translation model to automated cartoon coloring with very limited training data with industry applications in art, design and animation.

\section{A Small Data Problem - Automating Cartoon Coloring}

We consider the problem of automating the consistent coloring of a cartoon character type, in a flat color style like old Disney cel animation or The Simpsons. Is animated character coloring a small data problem? In many cases it's not, but sometimes, it is. 

If we were actually trying to color Bart Simpson or Mickey Mouse and had access to the Fox or Disney film archives, we might well have sufficient data for standard machine learning methods or at least enough footage to make it. There are 625 Simpsons episodes, nearly 20 million frames \citep{wikipedia:17}. Even for relatively rare characters in film or video, we might be able to leverage frame by frame continuity to color subsequent or intermediate frames, and when choosing production methods for new projects we could model 3D CG characters and color each only once using a 3D shader that imitates old style 2D cel coloring. 

Here, we explicitly want to study a case of scarcity, when a large body of work doesn't exist. When we have just a few dozen drawings, unconnected across time, only one of them colored, what can we do then?    

If we are to get from such a small data set to a machine learning model that can consistently paint the rest of the training drawings and the same artist's future drawings of the same type(s), we'll need extreme augmentation without overfitting. 

\section{The Geometry of Dragons: a Rule Based Solution for 80 Percent}

Faced with a shortage of training data, we should be asking if there is a good non machine learning based approach to our problem. If there's not a complete solution, is there a partial solution and would a partial solution do us any good? Do we even need machine learning to color flowers and dragons or can we specify geometric rules for coloring? 

\subsection{Tell a Kid How to Color}

\begin{figure}
   \centering
   \begin{subfigure}[b]{0.15\linewidth}
      \includegraphics[width=\linewidth]{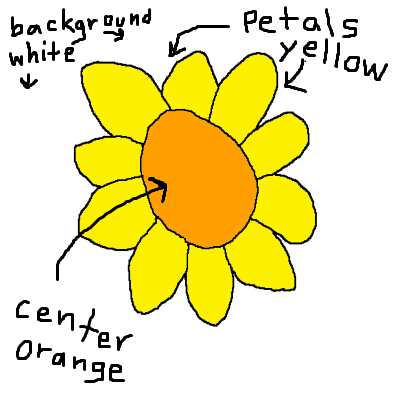}
   \end{subfigure}
   \begin{subfigure}[b]{0.15\linewidth}
      \includegraphics[width=\linewidth]{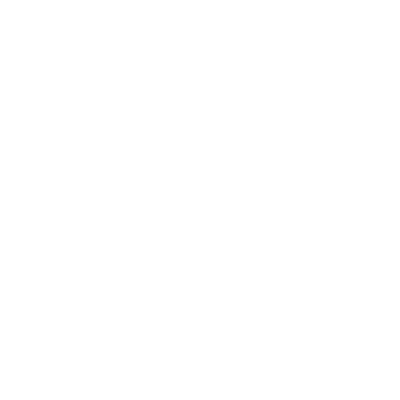}
   \end{subfigure}
   \begin{subfigure}[b]{0.15\linewidth}
      \includegraphics[width=\linewidth]{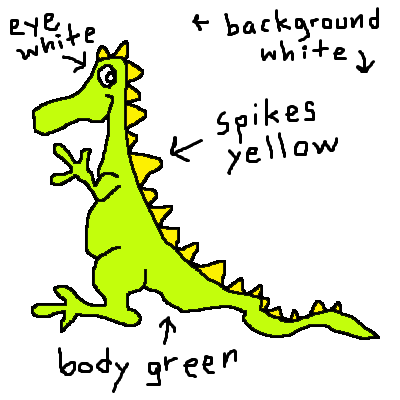}
   \end{subfigure}  
   \caption{Kid's rules: ``Color by body part.''}
   \label{fig:kidrulesflowerpart}
\end{figure}

When you don't know what rules to tell a computer, one way to start is to ask what rules you'd tell a person. If you were telling a kid how to color a flower or dragon, what would you tell them? You could probably give them one that is colored and just say, ``color the others like this.'' But to be more explicit about what ``like this'' means, you could state rules in terms of body or flower parts: Make the center orange. Make the petals yellow. Make the body green. Make the spikes yellow. Make the eyes white (Fig. 1). 

For flowers or simple dragons, we'd be done with just two or three rules. For fancy dragons, the rule list would be longer but still doable. We could write down a dozen ``color by body part'' rules to color wings, ears, belly scales, arms, legs, claws, etc. We'd tell the kid the extra rules or provide a couple of examples with the extra body parts and we'd expect our flowers and dragons would get colored the way we wanted them. 

But for a computer that doesn't know what a ``body'' or a ``petal'' is, it seems like we're as far away as ever. We could train a model to recognize body and flower parts but we'd need a huge amount of labeled training data which we don't have.

\subsection{From Body Parts to Geometric Rules}

What we'd like to do is directly translate our body part rules into geometric rules that are easy to program. The original coloring task was inspired by examples colored with a ``paint bucket'' tool which colors all the pixels in the same connected component as the clicked pixel. If we find the connected components in a drawing (which is done for us in existing libraries in e.g. Python), we can write down geometric rules using area and distance to label the parts of flowers (Fig. 2) and simple dragons and then use our original kids' color by body part rules. These geometric labeling rules don't work for all of our initial drawings but they give us a solid partial solution that works for about 80 percent. 

\paragraph{Rules for flowers (or simple dragons):} 
\begin{enumerate}
\item ``background''= biggest white component
\item ``center'' (or ``body'') = second biggest white component
\item ``petals'' (or ``spikes+'') = the remaining white components
\item (For dragons, add ``eye'' vs. ``spikes'' rule using distance to background)
\end{enumerate}

For at least some of the fancy dragon features, we can keep going, finding geometric rules that distinguish between claws and spikes or between eyes and eyelids. But even if we had geometric rules for every feature, our basic geometric rules for labeling body parts are fundamentally flawed. They only work for most of the drawings. For every rule I've written, I have a drawing it doesn't work for. Where do the rules break down? Why don't they work for all drawings? And if they aren't reliable, what good are they?

\begin{figure}
   \centering
   \begin{subfigure}[b]{0.15\linewidth}
      \includegraphics[width=\linewidth]{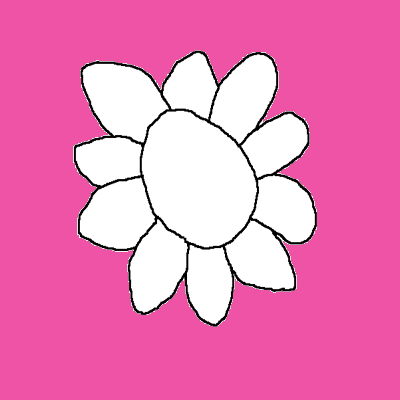}
      \caption{Not flower}
   \end{subfigure}
   \begin{subfigure}[b]{0.15\linewidth}
      \includegraphics[width=\linewidth]{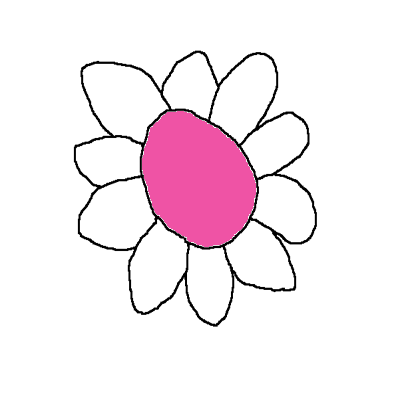}
      \caption{Center}
   \end{subfigure}
   \begin{subfigure}[b]{0.15\linewidth}
      \includegraphics[width=\linewidth]{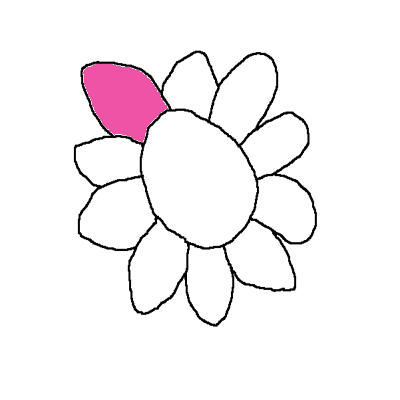}
      \caption{One petal}
   \end{subfigure} 
   \begin{subfigure}[b]{0.15\linewidth}
      \includegraphics[width=\linewidth]{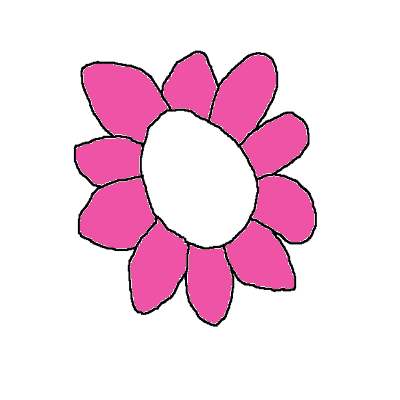}
      \caption{All petals}
   \end{subfigure}  
   \caption{Color flower by component}
   \label{fig:colorbycomp}
\end{figure}

\section{From 80 Percent to a Full Training Set With Rule Breaking Transformations and Extreme Augmentation}
Our geometric rules work to color about 80 percent of the original drawings. We'll call those drawings ``rule conforming'', write a program to color them and make those AB pairs the beginning of our training set. Then what?

\subsection{Where the Rules Break Down - the Other 20 Percent}

Let's examine the question of where the rules break down (Fig. 3). Writing geometric rules, we are formalizing assumptions that aren't true for all of our drawings: The background is connected. The background is the biggest component. Flower centers are bigger than all the petals. There are no gaps in drawn lines. Body and limbs are connected.  

\begin{figure}
   \centering
   \begin{subfigure}[b]{0.15\linewidth}
      \includegraphics[width=\linewidth]{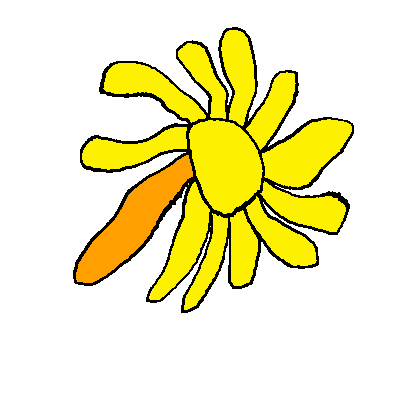}
   \end{subfigure} 
   \begin{subfigure}[b]{0.15\linewidth}
      \includegraphics[width=\linewidth]{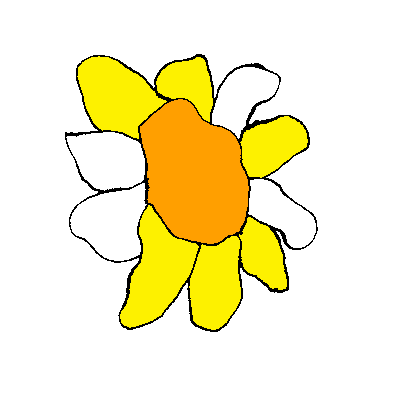}
   \end{subfigure}
   \begin{subfigure}[b]{0.15\linewidth}
      \includegraphics[width=\linewidth]{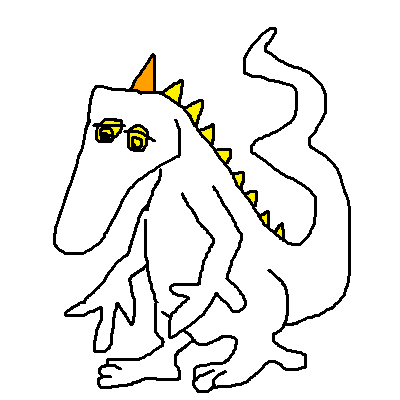}
   \end{subfigure}      
   \begin{subfigure}[b]{0.15\linewidth}
      \includegraphics[width=\linewidth]{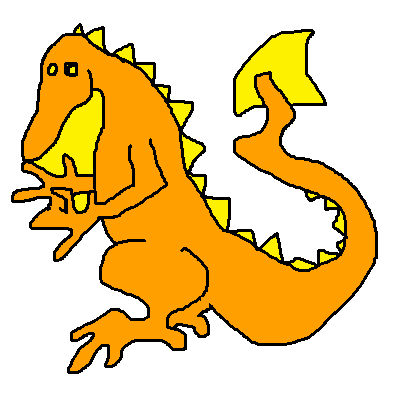}
   \end{subfigure}   
   \begin{subfigure}[b]{0.15\linewidth}
      \includegraphics[width=\linewidth]{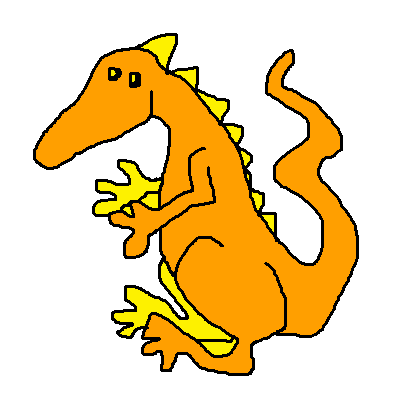}
   \end{subfigure} 
   \caption{Where rules break down: line gaps, sizes, back limbs}
   \label{fig:breakrules}
\end{figure}

It's these cases that we hope to train our model to take care of because our geometric rules don't. But because they couldn't be rule colored, they aren't in our training set. We need to find a way to add them.  

\subsection{Rule Breaking Transfomations}
For each of the assumptions/rules we made, we look for rule breaking transformations that we can apply to a rule conforming drawing A or to a drawing/colored AB pair (by applying the same transformation to A and to B) so that we can augment our AB training set with pairs that break all the rules. We use domain knowledge where appropriate to choose the best functions and parameters.

\textit{Background is connected and is the biggest component.} We cropped, scaled or translated to disconnect and/or shrink the background.
 
\textit{Center of flower smaller than petals.} We assumed our flowers all looked like a sunflower with a big center and smaller petals but what about the ones that look more like a daisy with a small center and big petals or petals of various sizes? 

We'd like a transformation that can turn our sunflowers into daisies. Since the flowers have an approximate radial symmetry, we switched to polar coordinates, scaled to embed our image square in a unit disk and looked at radially symmetric homeomorphisms of the disk to find a function which would disproportionately shrink the center of our image and make a synthetic daisy. $f(r,\theta)=(r^3,\theta)$ worked perfectly to create daisies (Fig. 4) but distorted dragons well beyond natural variation and almost beyond recognition.

\begin{figure}
   \centering
   \begin{subfigure}[b]{0.15\linewidth}
      \includegraphics[width=\linewidth]{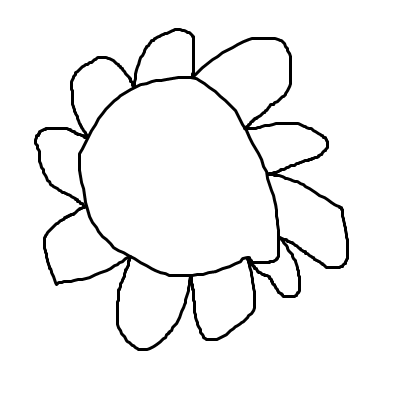}
   \end{subfigure}
   \begin{subfigure}[b]{0.15\linewidth}
      \includegraphics[width=\linewidth]{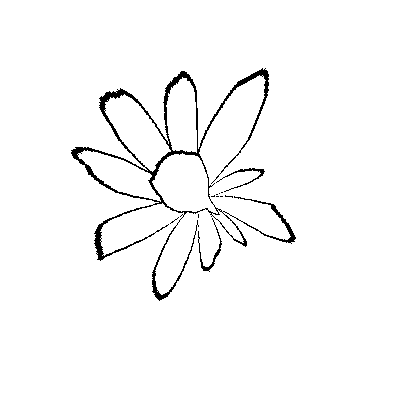}
   \end{subfigure}  
   \begin{subfigure}[b]{0.15\linewidth}
      \includegraphics[width=\linewidth]{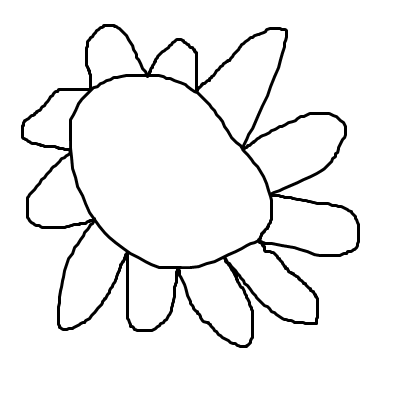}
   \end{subfigure}
   \begin{subfigure}[b]{0.15\linewidth}
      \includegraphics[width=\linewidth]{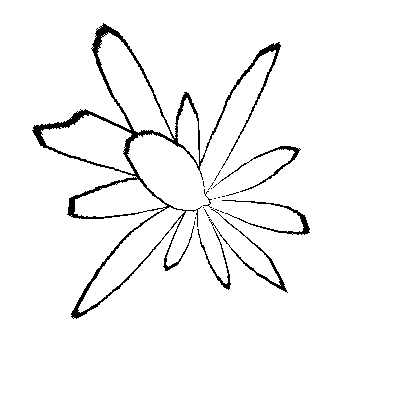}
   \end{subfigure} 
   \caption{Synthetic daisy creation with $f(r,\theta)=(r^3,\theta)$}
   \label{fig:rtorcubed}
\end{figure}

\textit{Gaps in lines.}
We expect some test drawings will have poorly connected lines where there's a gap; where, for example, the petal doesn't quite connect to the center in a hastily drawn line. To create gaps in lines in training data, we made erasures on a drawing A from an AB pair, either by hand or in an automated way using a squiggle as a mask, and then paired the new A$'$ with its old colored version B to get a new A$'$B pair with line gaps for the training set. Other functions occasionally created gaps by stretching/thinning lines.

\subsection{Additional Augmentation}

Having filled in the training set with many of the kinds of data we knew we were missing, what else can we do to add useful variation to the data while avoiding overfitting?

\textit{Affine transformations.}
We included the usual affine transformation augmentations: translation, rotation, scale, skew and mirror flip. \citep[See][]{Bloice:17}.

\textit{Elastic distortions/Gaussian blur.} 
We extended an idea from \citet{simard:03} to use Gaussian blur to change the drawn line. Simard et al. used it to augment the MNIST OCR training set, comparing it to the natural oscillations of the hand. \citep[For an implementation see ][]{ernie:17}. A very useful transformation type, elastic distortions preserved pose and major features but changed characters in interesting and believable ways. For example, two distortions gave the same dragon either a fat, short snout or a long, pointy one and changed the bend in its tail (Fig. 5).

\textit{Composition of functions.}
For each original AB pair, we composed multiple sequences of transformations to get new AB pairs. Two composition dangers to watch out for are half dragons in space (cropping an edge and then moving that cut edge back into the middle) and unintended duplicates (beware of commutativity.)

\section{Identify Target ML Model and Training Needs}

As a general problem solving strategy, when the outline of our proposed solution has multiple parts, we should have some level of confidence that we'll be able to solve the other parts before we invest too much effort into one. To the extent that our approach to one part affects another, we should go beyond likelihood estimates and have an idea what the subpart solutions might look like. 

Before beginning our quest for rule breaking transformations and extreme augmentation, we should have had an idea of the kind of machine learning model that should work if we could build a training set and what kind and how much training data we'd likely need. Our target number for a deep learning model might have been in the tens of thousands or hundreds of thousands. It was only the identification of a suitable existing model with comparatively modest data needs that made sufficient augmentation seem possible.

\begin{figure}
   \centering
   \begin{subfigure}[b]{0.15\linewidth}
      \includegraphics[width=\linewidth]{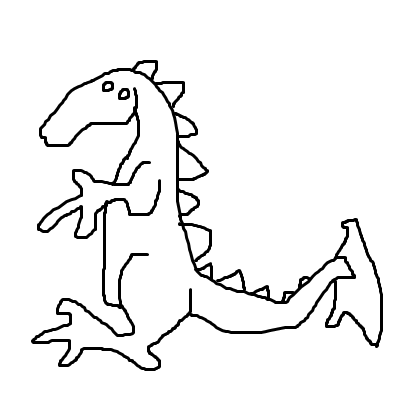}
   \end{subfigure}
   \begin{subfigure}[b]{0.15\linewidth}
      \includegraphics[width=\linewidth]{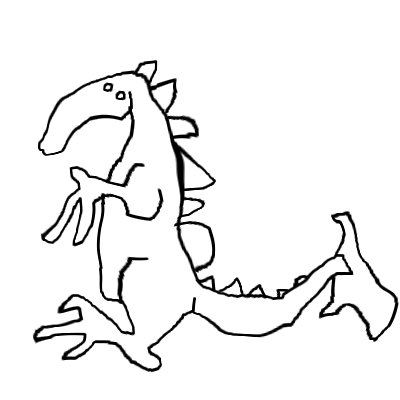}
   \end{subfigure}  
   \begin{subfigure}[b]{0.15\linewidth}
      \includegraphics[width=\linewidth]{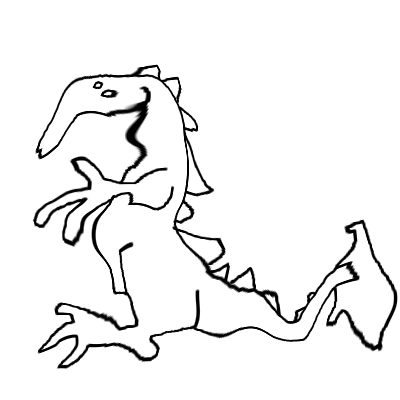}
   \end{subfigure}
   \begin{subfigure}[b]{0.15\linewidth}
      \includegraphics[width=\linewidth]{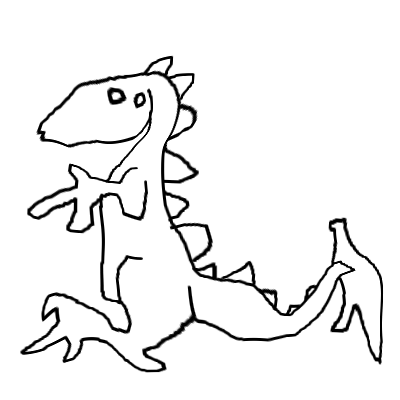}
   \end{subfigure}
   \begin{subfigure}[b]{0.15\linewidth}
      \includegraphics[width=\linewidth]{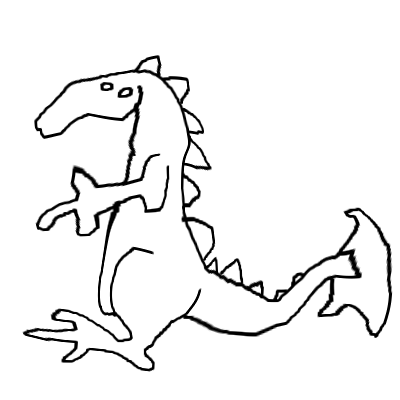}
   \end{subfigure}        
   \caption{Elastic distortion/Gaussian blur}
   \label{fig:elastic}
\end{figure}

We have a natural pairing of images in two styles (uncolored drawing and its colored version) and we want to go from one of the first kind to one of the second kind, which made our cartoon coloring application look like an excellent candidate for applying ``Image-to-Image Translations with Conditional Adversarial Nets'' \citep{pix2pix:16}. Their facades and city maps applications needed only 400-1100 AB training pairs. So we made that our target range. That's still an ambitious order of magnitude more than our original drawing data and two orders of magnitude more than our hand colored data, but far less than we might have expected.

\section{Experiments, Results and Future Research} 

We had excellent results with uncropped flowers and promising results with dragons from only a few dozen original drawings colored and augmented as described above (all trained with orange/yellow scheme). The models handled several issues our geometric rules could not - flower line gaps, small centers and coloring dragon back limbs but fell short of our aspirations on croppings, background areas near close parts, all yellow spikes and fancy dragon parts.

\begin{figure}
   \centering
   \begin{subfigure}[b]{0.05\linewidth}
      \includegraphics[width=\linewidth]{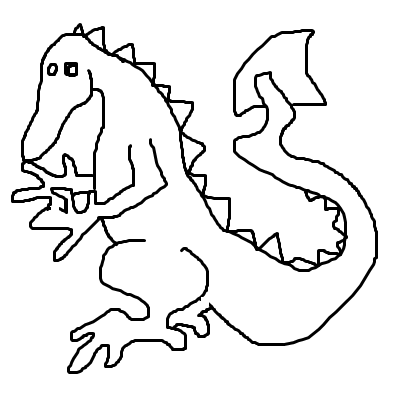}
   \end{subfigure} 
   \begin{subfigure}[b]{0.05\linewidth}
      \includegraphics[width=\linewidth]{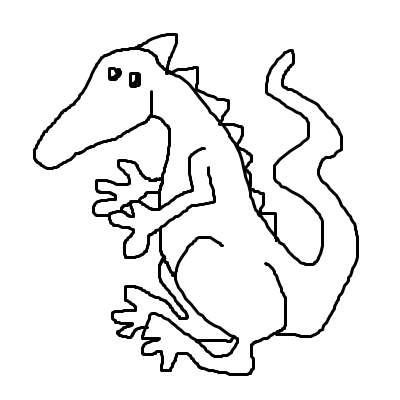}
   \end{subfigure} 
   \begin{subfigure}[b]{0.05\linewidth}
      \includegraphics[width=\linewidth]{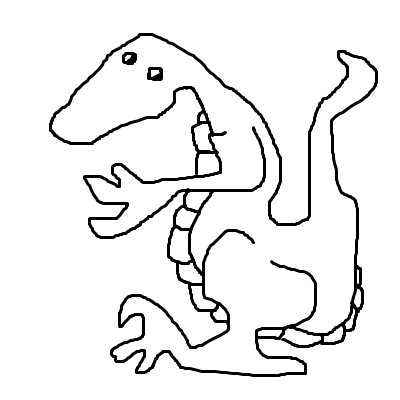}
   \end{subfigure} 
   \begin{subfigure}[b]{0.05\linewidth}
      \includegraphics[width=\linewidth]{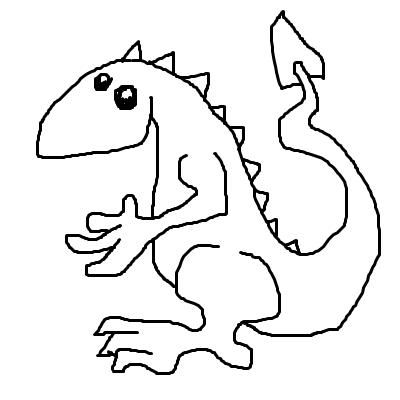}
   \end{subfigure} 
   \begin{subfigure}[b]{0.05\linewidth}
      \includegraphics[width=\linewidth]{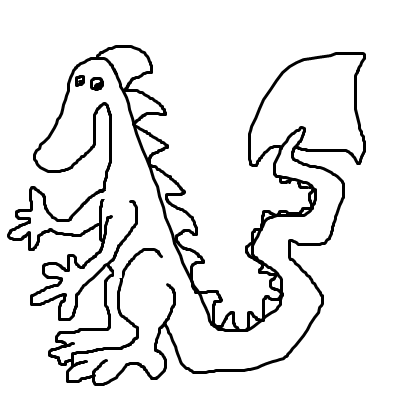}
   \end{subfigure} 
   \begin{subfigure}[b]{0.05\linewidth}
      \includegraphics[width=\linewidth]{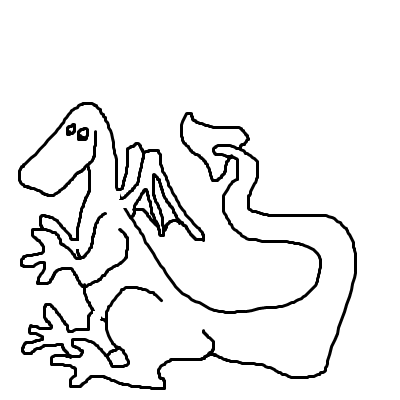}
   \end{subfigure} 
   \begin{subfigure}[b]{0.05\linewidth}
      \includegraphics[width=\linewidth]{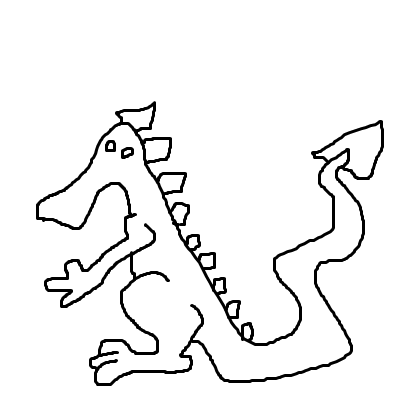}
   \end{subfigure} 
   \begin{subfigure}[b]{0.05\linewidth}
      \includegraphics[width=\linewidth]{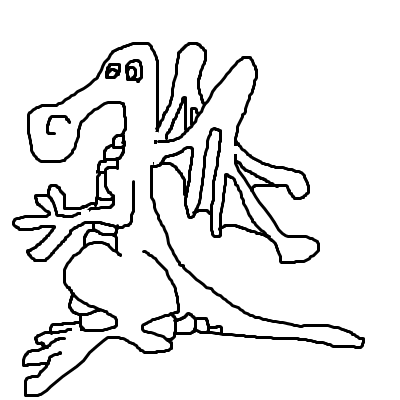}
   \end{subfigure} 
   \begin{subfigure}[b]{0.05\linewidth}
      \includegraphics[width=\linewidth]{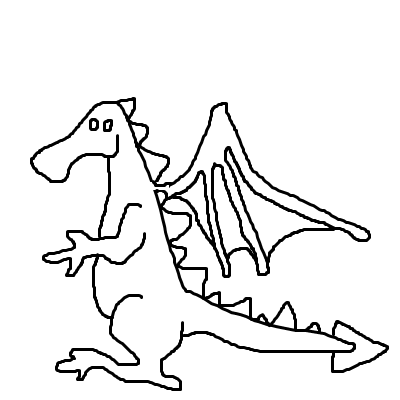}
   \end{subfigure} 
   \begin{subfigure}[b]{0.05\linewidth}
      \includegraphics[width=\linewidth]{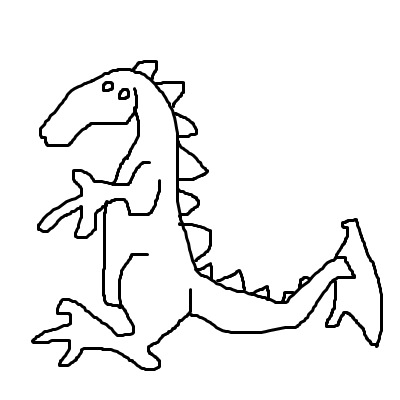}
   \end{subfigure} 
   \begin{subfigure}[b]{0.05\linewidth}
      \includegraphics[width=\linewidth]{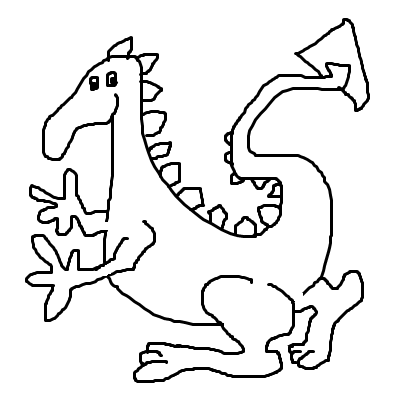}
   \end{subfigure} 
   \begin{subfigure}[b]{0.05\linewidth}
      \includegraphics[width=\linewidth]{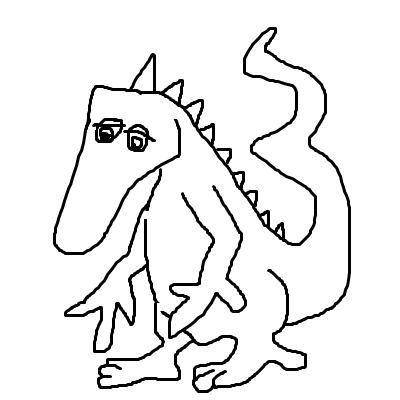}
   \end{subfigure} 
   \begin{subfigure}[b]{0.05\linewidth}
      \includegraphics[width=\linewidth]{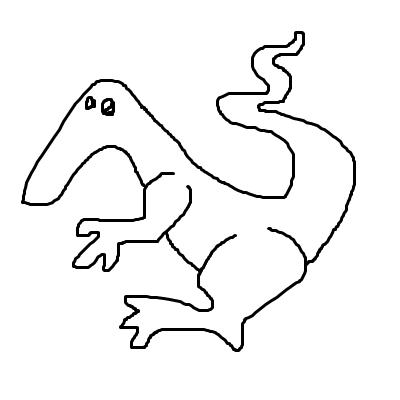}
   \end{subfigure} 
   \begin{subfigure}[b]{0.05\linewidth}
      \includegraphics[width=\linewidth]{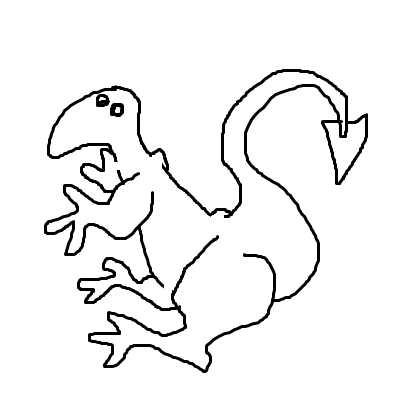}
   \end{subfigure} 
   \begin{subfigure}[b]{0.05\linewidth}
      \includegraphics[width=\linewidth]{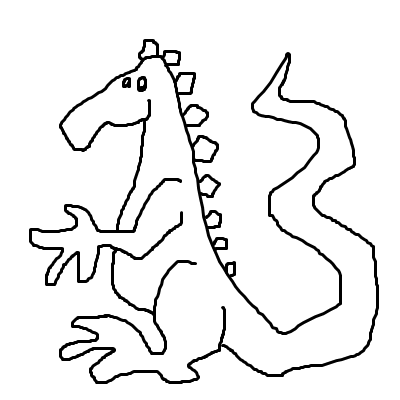}
   \end{subfigure} 
   \begin{subfigure}[b]{0.05\linewidth}
      \includegraphics[width=\linewidth]{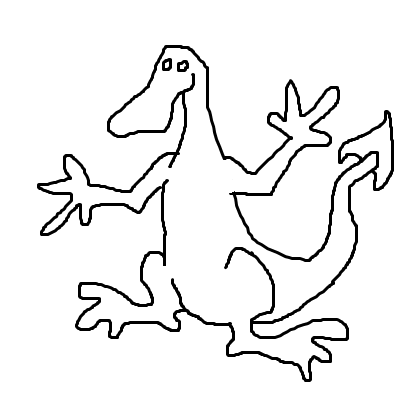}
   \end{subfigure} 
   \begin{subfigure}[b]{0.05\linewidth}
      \includegraphics[width=\linewidth]{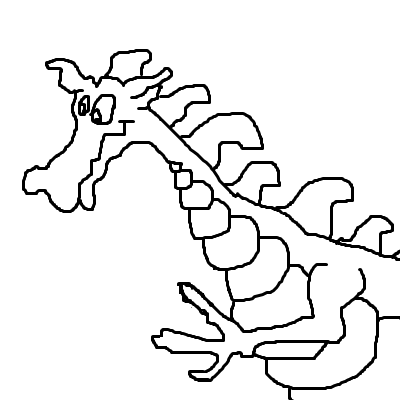}
   \end{subfigure} 
   \begin{subfigure}[b]{0.05\linewidth}
      \includegraphics[width=\linewidth]{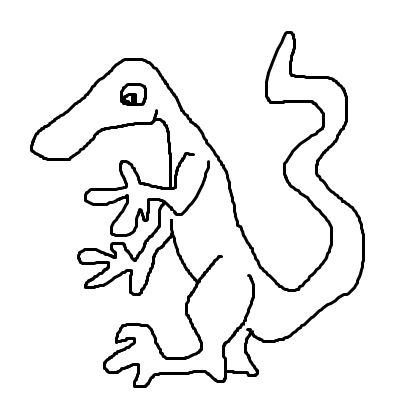}
   \end{subfigure}
   \begin{subfigure}[b]{0.05\linewidth}
      \includegraphics[width=\linewidth]{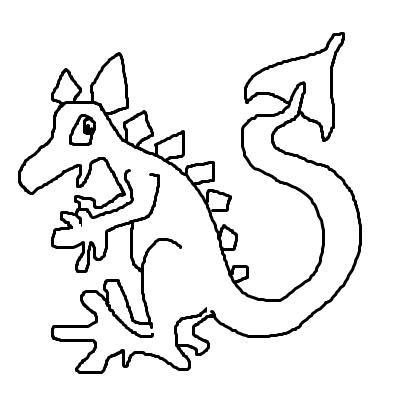}
   \end{subfigure} 
   \begin{subfigure}[b]{0.05\linewidth}
      \includegraphics[width=\linewidth]{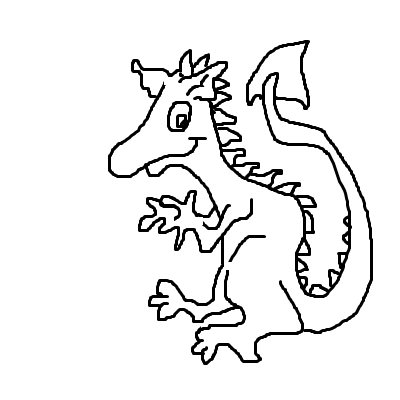}
   \end{subfigure}
   \begin{subfigure}[b]{0.05\linewidth}
      \includegraphics[width=\linewidth]{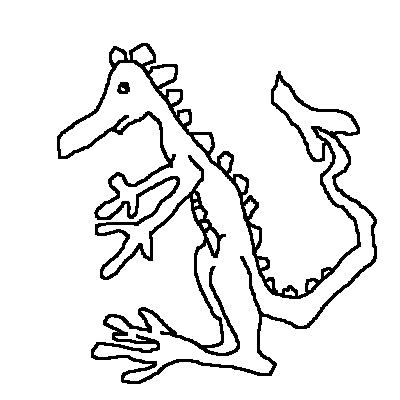}
   \end{subfigure} 
   \begin{subfigure}[b]{0.05\linewidth}
      \includegraphics[width=\linewidth]{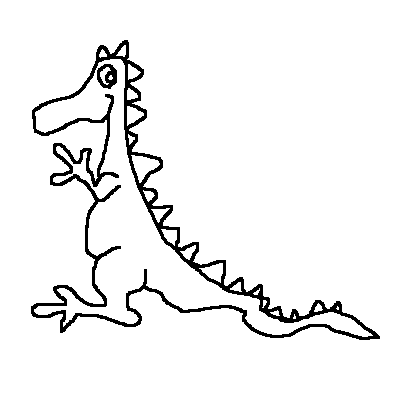}
   \end{subfigure} 
   \begin{subfigure}[b]{0.05\linewidth}
      \includegraphics[width=\linewidth]{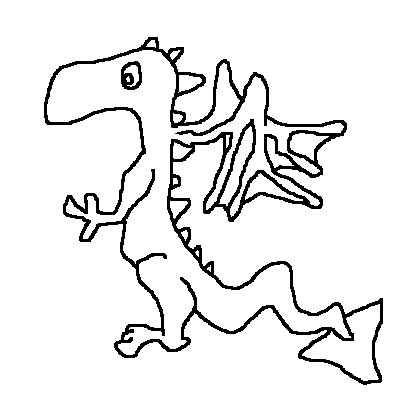}
   \end{subfigure} 
   \begin{subfigure}[b]{0.05\linewidth}
      \includegraphics[width=\linewidth]{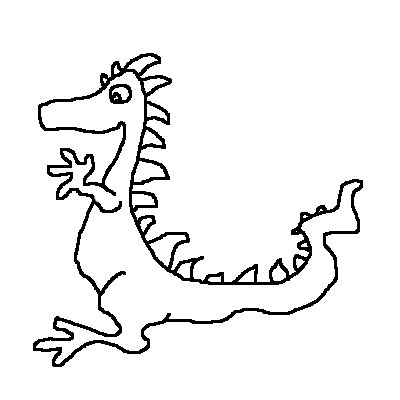}
   \end{subfigure} 
   \begin{subfigure}[b]{0.05\linewidth}
      \includegraphics[width=\linewidth]{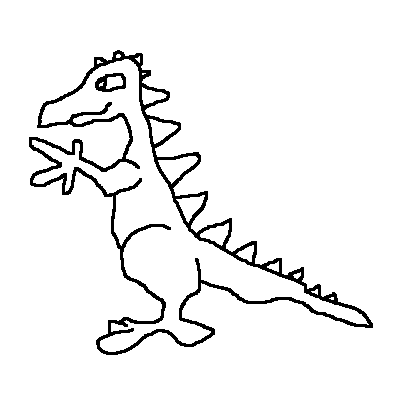}
   \end{subfigure} 
   \begin{subfigure}[b]{0.05\linewidth}
      \includegraphics[width=\linewidth]{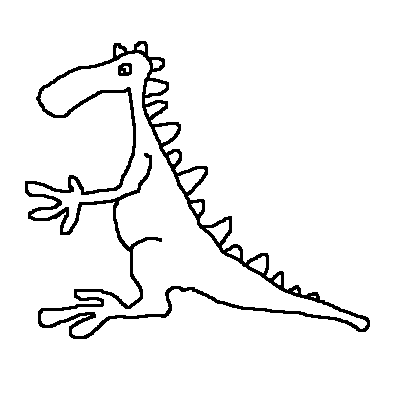}
   \end{subfigure} 
   \begin{subfigure}[b]{0.05\linewidth}
      \includegraphics[width=\linewidth]{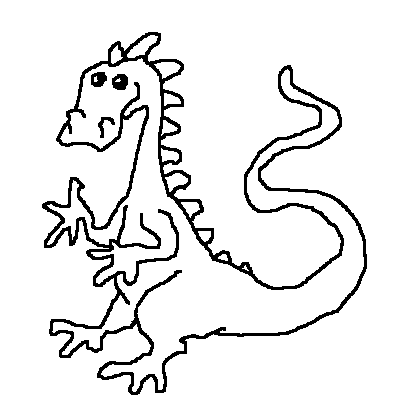}
   \end{subfigure} 
   \begin{subfigure}[b]{0.05\linewidth}
      \includegraphics[width=\linewidth]{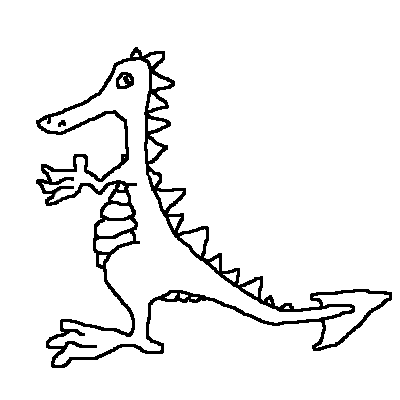}
   \end{subfigure}
   \begin{subfigure}[b]{0.05\linewidth}
      \includegraphics[width=\linewidth]{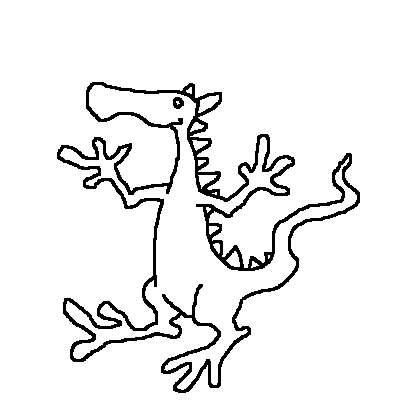}
   \end{subfigure} 
   \begin{subfigure}[b]{0.05\linewidth}
      \includegraphics[width=\linewidth]{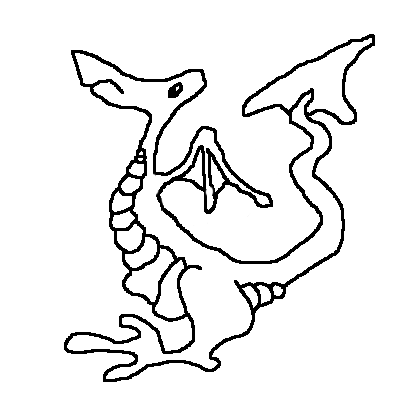}
   \end{subfigure}
   \begin{subfigure}[b]{0.05\linewidth}
      \includegraphics[width=\linewidth]{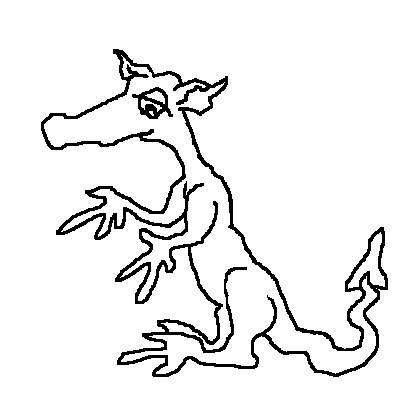}
   \end{subfigure} 
   \begin{subfigure}[b]{0.05\linewidth}
      \includegraphics[width=\linewidth]{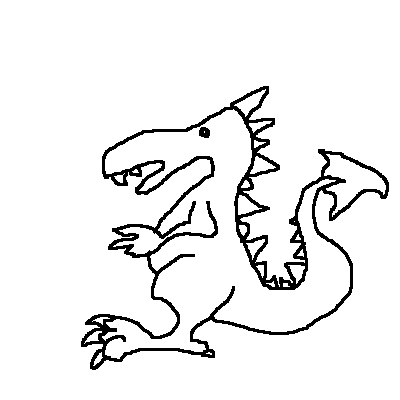}
   \end{subfigure}
   \caption{Every hand drawn picture that was used in making the 32Dragons training set}
   \label{fig:32DragonsOriginalData}
\end{figure}

We ran three experiments on a single AWS GPU instance using Hesse's TensorFlow \citep{hesse:17} implementation of Isola et al.'s Pix2Pix image translation model. We trained on all flowers, all dragons and both dragons and flowers for 200 epochs. We then tested each trained model on a mixed test set of new flowers, dragons and a few others.

Our original 40 uncolored ``flower'' and 32 uncolored ``dragon'' (Fig. 6) 400x400 px drawings were made in a standard computer Paint program by the author. We trained for an orange/yellow coloring scheme for both flowers and dragons to see if the similarity between flower centers/petals and dragon bodies/spikes would make a mixed training set useful for dragons.  

\subsection{Experiment 1: 40Flowers} 
\textit{Training data: }40 original flower drawings, rule colored and augmented, including erasures to 506 AB training pairs. 

\textit{Results: }40Flowers was very good at coloring uncropped flowers. It could handle all the flowers our geometric rules could and several types they could not - line gaps, small centers and various size petals. With dragons, strongly cropped flowers and ``other'', it didn't color according to our intended scheme but it often (but not always) recognized lines as color boundaries (Fig. 7).

\textit{Next steps: }The relative simplicity of the character type and the promising family of homeomorphisms of the disk suggest we could successfully shrink the original flower training set even further and/or improve cropping performance.

\begin{figure}
   \centering
   \begin{subfigure}[b]{0.1\linewidth}
      \includegraphics[width=\linewidth]{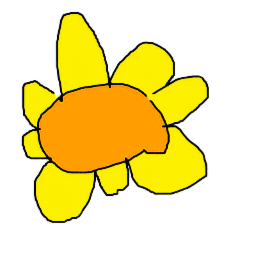}
   \end{subfigure} 
   \begin{subfigure}[b]{0.1\linewidth}
      \includegraphics[width=\linewidth]{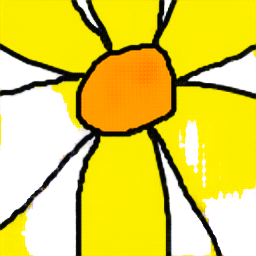}
   \end{subfigure}    
   \begin{subfigure}[b]{0.1\linewidth}
      \includegraphics[width=\linewidth]{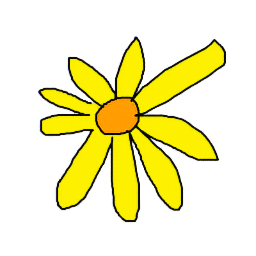}
   \end{subfigure}   
   \begin{subfigure}[b]{0.1\linewidth}
      \includegraphics[width=\linewidth]{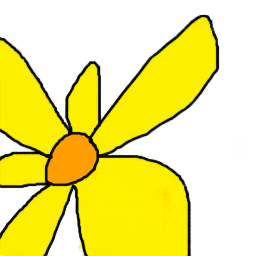}
   \end{subfigure}  
   \begin{subfigure}[b]{0.1\linewidth}
      \includegraphics[width=\linewidth]{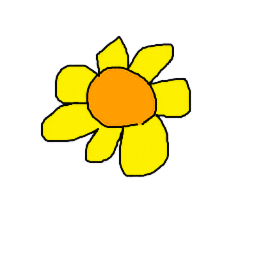}
   \end{subfigure}  
   \begin{subfigure}[b]{0.1\linewidth}
      \includegraphics[width=\linewidth]{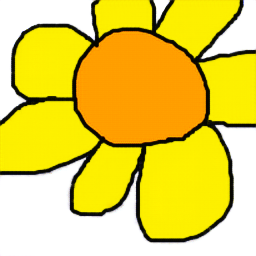}
   \end{subfigure} 
   \begin{subfigure}[b]{0.1\linewidth}
      \includegraphics[width=\linewidth]{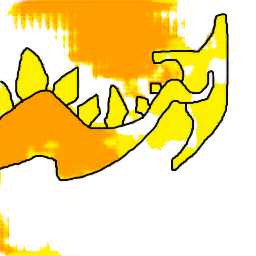}
   \end{subfigure}
   \begin{subfigure}[b]{0.1\linewidth}
      \includegraphics[width=\linewidth]{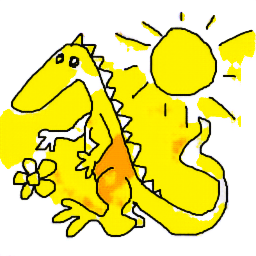}
   \end{subfigure}  
   \begin{subfigure}[b]{0.1\linewidth}
      \includegraphics[width=\linewidth]{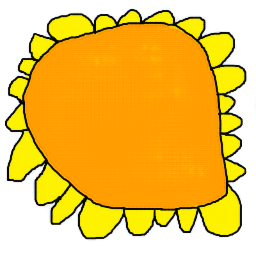}
   \end{subfigure}
   \caption{Colored by model 40Flowers (trained only on flowers)}
   \label{fig:40FlowersTest}
\end{figure}
\subsection{Experiment 2: 32Dragons} 

\textit{Training data:} 32 original dragon drawings, rule colored and augmented to 469 AB training pairs. We added hand erasures or ``paint bucket'' fills to a half dozen colored drawings to add properly colored back limbs to the training set. 

\textit{Results: }32Dragons had quite good results on simple dragons, but often colored a few spikes orange and muddled orange and yellow at the tail tip or tail spike, possibly confused by orange ears and the inconsistency of a yellow tail spike's existence in the training set. It did well with the simpler eye style and struggled with the other. It colored background where the gap between body parts was small. It did a lot right with wings, back limbs and ``other'' pictures but didn't color them quite according to our scheme. It struggled a bit with gaps, going a bit outside the line gap and introducing unwanted black in other places. 32Dragons colored flowers in an unexpected way with mostly white centers and petals mostly orange with white and yellow mixed in. There was often some color in the background. Croppings were worse. Dragon eyes may explain the coloring of flowers since the the round eye is white surrounded by orange body (Fig. 8).

\textit{Next steps: }Several research directions seem promising. We could use transfer learning to build from simpler to more complex characters (both across character types or within a single type adding body part features) and to incorporate problematic transformations (e.g erasures or croppings.) We could augment a purpose built data set by saving in progress drawings at multiple stages. We could look for additional distortion transformations and investigate methods for drawing automation. Relevant works include \citet{Hauberg:16} for learned diffeomorphisms and \citet{Ha:17} for sketching with neural nets. 
 
\begin{figure}
   \centering
   \begin{subfigure}[b]{0.1\linewidth}
      \includegraphics[width=\linewidth]{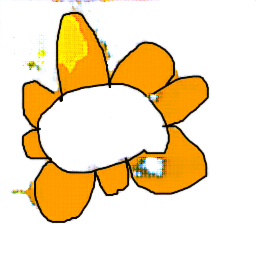}
   \end{subfigure}
   \begin{subfigure}[b]{0.1\linewidth}
      \includegraphics[width=\linewidth]{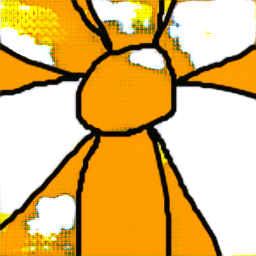}
   \end{subfigure}
   \begin{subfigure}[b]{0.1\linewidth}
      \includegraphics[width=\linewidth]{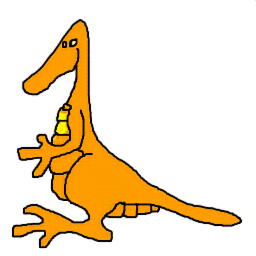}
   \end{subfigure}  
   \begin{subfigure}[b]{0.1\linewidth}
      \includegraphics[width=\linewidth]{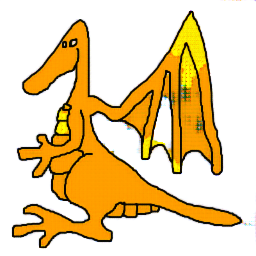}
   \end{subfigure}
   \begin{subfigure}[b]{0.1\linewidth}
      \includegraphics[width=\linewidth]{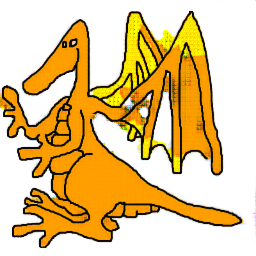}
   \end{subfigure}
   \begin{subfigure}[b]{0.1\linewidth}
      \includegraphics[width=\linewidth]{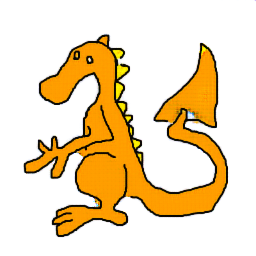}
   \end{subfigure}
   \begin{subfigure}[b]{0.1\linewidth}
      \includegraphics[width=\linewidth]{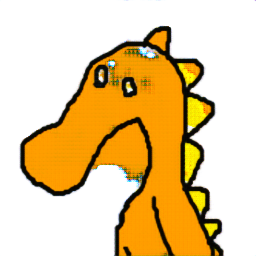}
   \end{subfigure} 
   \begin{subfigure}[b]{0.1\linewidth}
      \includegraphics[width=\linewidth]{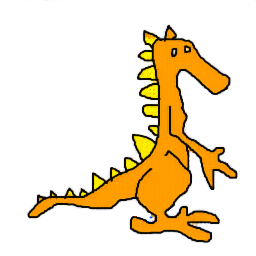}
   \end{subfigure}
   \begin{subfigure}[b]{0.1\linewidth}
      \includegraphics[width=\linewidth]{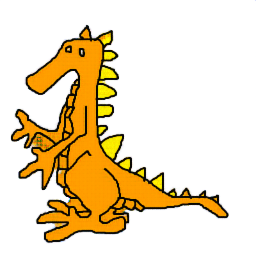}
   \end{subfigure}
   \begin{subfigure}[b]{0.1\linewidth}
      \includegraphics[width=\linewidth]{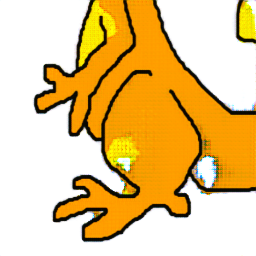}
   \end{subfigure}
   \begin{subfigure}[b]{0.1\linewidth}
      \includegraphics[width=\linewidth]{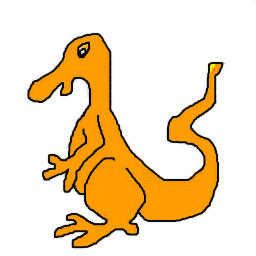}
   \end{subfigure}
   \begin{subfigure}[b]{0.1\linewidth}
      \includegraphics[width=\linewidth]{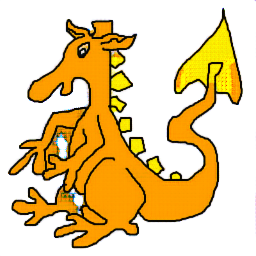}
   \end{subfigure}
   \begin{subfigure}[b]{0.1\linewidth}
      \includegraphics[width=\linewidth]{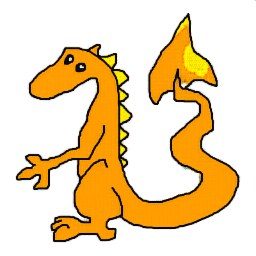}
   \end{subfigure} 
   \begin{subfigure}[b]{0.1\linewidth}
      \includegraphics[width=\linewidth]{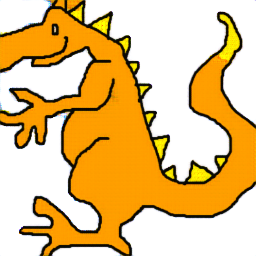}
   \end{subfigure}  
   \begin{subfigure}[b]{0.1\linewidth}
      \includegraphics[width=\linewidth]{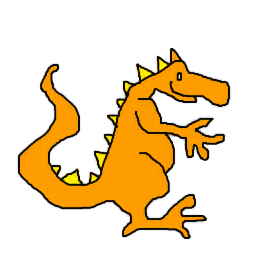}
   \end{subfigure}
   \begin{subfigure}[b]{0.1\linewidth}
      \includegraphics[width=\linewidth]{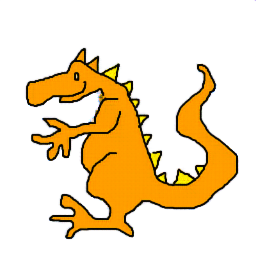}
   \end{subfigure}         
   \begin{subfigure}[b]{0.1\linewidth}
      \includegraphics[width=\linewidth]{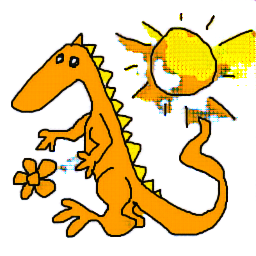}
   \end{subfigure}
   \begin{subfigure}[b]{0.1\linewidth}
      \includegraphics[width=\linewidth]{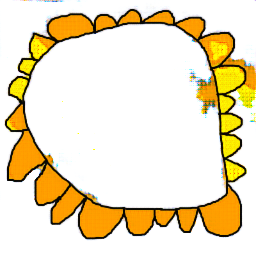}
   \end{subfigure}  
   \caption{Colored by model 32Dragons (trained only on dragons)}
   \label{fig:32DragonsTest}
\end{figure}

\subsection{Experiment 3: MixedFD} 

\textit{Training Data: }Combined 40Flowers and 32Dragons training sets 975 AB training pairs. 

\textit{Results: }MixedFD had quite good results on simple dragons, but not as good as 32Dragons. MixedFD miscolored the eyes, probably confused by orange flower centers. MixedFD colored spikes yellow, but at the expense of miscolored ears. MixedFD was quite good on dragon line gaps and it was a little better than 32Dragons on crops but it introduced even more unwanted black. MixedFD was good on uncropped flowers, including line gaps and various center/petal sizes, but had unwanted black.

\section{Industry Applications and Further Research}

This research is the first step in a broader inquiry into how to generate new works in the style of an individual artist from limited original data, with industry applications in art, design and animation. Automating the coloring of cartoon characters with a consistent color scheme has potential applications in cel style animation for film and game production and in comics for periodical and book publication, but our target application is more ambitious. 

In fashion, interior design and architecture, from high end photography printers to modern looms to CNC steel fabrication equipment, wherever there exists a manufacturing process where it is cheap to switch patterns, there is the possibility of offering products incorporating unique to each customer designs at mass market prices if there is an inexpensive way to create new designs. The challenge is to maintain design style and quality across a large number of items while introducing automation and variation. 

For a first use case in wearables, we are currently using the work in this paper together with further work on drawing creation to produce a family of designs for T-shirts where each shirt will have a unique but closely related design. Shirt mock ups shown here (Fig. 9) use test image results from the 32Dragons experiment. 

\begin{figure}
   \centering
   \begin{subfigure}[b]{0.1\linewidth}
      \includegraphics[width=\linewidth]{PicsForPAPIsPaper/Dragon1mirror-outputs.png}
   \end{subfigure}
   \begin{subfigure}[b]{0.15\linewidth}
      \includegraphics[width=\linewidth]{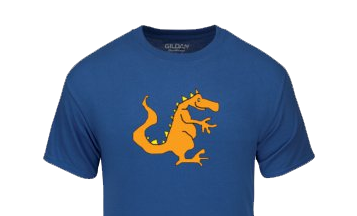}
   \end{subfigure}  
   \begin{subfigure}[b]{0.1\linewidth}
      \includegraphics[width=\linewidth]{PicsForPAPIsPaper/Dragon2-outputs.png}
   \end{subfigure}
   \begin{subfigure}[b]{0.15\linewidth}
      \includegraphics[width=\linewidth]{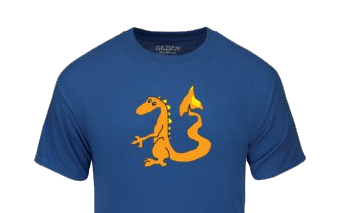}
   \end{subfigure}        
   \caption{AI T-shirt designs: unique design for every customer}
   \label{fig:DragonShirts}
\end{figure}

\section{Conclusion}

We are exploring ways to use machine learning for automated design generation, with applications across multiple industries. To automate cartoon coloring, we suggested a generalizable strategy for going from a very small data set to a deep learning training set, bootstrapping with geometric rules to get to a partial solution and using strategic rulebreaking transformations and other augmentations to go the rest of the way.  

The study of a specific application has allowed for a useful discussion of implementation details and specific outcomes, but the underlying ideas generalize beyond the problem of coloring dragons. Up a few levels of abstraction, the strategy's formulation is equally applicable to image, language or numerical data across a wide range of domains. Given any unlabeled data set, if there is a substantial subset with a method for automated labeling, a method for transforming that labeled subset into labeled data like the data we couldn't automate labeling for and other application appropriate, label preserving augmentation techniques (if the original data set was small), then we can combine these methods to create a large labeled training set representing the variation in our original small unlabeled training set (and maybe well beyond). 

\vskip 0.2in
\bibliography{greenebibBoston2017}

\end{document}